\documentclass[runningheads]{llncs}

\usepackage{eccv}

\usepackage{eccvabbrv}

\usepackage{graphicx}
\usepackage{booktabs}

\usepackage[ruled, linesnumbered, vlined]{algorithm2e}
\usepackage{multirow}
\usepackage{array}
\usepackage{tabularray}
\newcolumntype{P}[1]{>{\centering\arraybackslash}p{#1}}

\usepackage[accsupp]{axessibility}  

\usepackage{hyperref}

\begin{document}

\title{Tracking-Assisted Object Detection with Event Cameras} 


\author{Ting-Kang Yen\inst{1} \and
Igor Morawski\inst{1} \and
Shusil Dangi\inst{3} \and
Kai He\inst{3} \and
Chung-Yi Lin\inst{1} \and
Jia-Fong Yeh\inst{1} \and
Hung-Ting Su\inst{1}\and
Winston Hsu\inst{1,2}}

\authorrunning{T.-K. Yen et al.}

\institute{National Taiwan University \and
Mobile Drive Technology \and
Qualcomm Inc.}

\maketitle

\begin{abstract}

Event-based object detection has recently garnered attention in the computer vision community due to the exceptional properties of event cameras, such as high dynamic range and no motion blur. However, feature asynchronism and sparsity cause \emph{invisible objects} due to no relative motion to the camera\footnotemark{}, posing a significant challenge in the task. Prior works have studied various implicit-learned memories to retain as many temporal cues as possible. However, implicit memories still struggle to preserve long-term features effectively.
In this paper, we consider those \emph{invisible objects} as \emph{pseudo-occluded objects} and aim to detect them by tracking through occlusions. Firstly, we introduce \emph{visibility} attribute of objects and contribute an auto-labeling algorithm to not only clean the existing event camera dataset but also append additional \emph{visibility} labels to it. Secondly, we exploit tracking strategies for pseudo-occluded objects to maintain their permanence and retain their bounding boxes, even when features have not been available for a very long time. These strategies can be treated as an explicit-learned memory guided by the tracking objective to record the displacements of objects across frames. Lastly, we propose a spatio-temporal feature aggregation module to enrich the latent features and a consistency loss to increase the robustness of the overall pipeline.
We conduct comprehensive experiments to verify our method's effectiveness where still objects are retained, but real occluded objects are discarded. The results demonstrate that (1) the additional \emph{visibility} labels can assist in supervised training, and (2) our method outperforms state-of-the-art approaches with a significant improvement of 7.9\% absolute mAP. Our code and the clean dataset with additional \emph{visibility} labels are available at \url{https://github.com/tkyen1110/TEDNet}.

\footnotetext{The objects with no relative motion to the camera, still objects, invisible objects, and pseudo-occluded objects are used interchangeably in the paper.}
\end{abstract}
\section{Introduction}
\label{sec:intro}
Event cameras measure only per-pixel intensity changes asynchronously and are suitable for detecting fast-moving objects. For those slow-moving objects or still objects, it is very hard to notice their physical existence if there are no temporal cues. For example, \cref{fig:probdef} shows that 
the features in the ground truth bounding boxes disappear gradually when slowing down, become empty when being still, and reappear gradually when speeding up. Human beings are capable of knowing the physical existence of these two cars as long as they observe the whole video sequence, even if there are no features in the ground truth bounding boxes. This kind of ability is known as object permanence\cite{op1985,op1991,op2015}. In fact, object permanence is a term from developmental psychology that describes how infants know the continuous existence of unseen objects.

\newlength{\textwidthA}
\newlength{\textwidthB}
\setlength{\textwidthA}{0.16\textwidth}
\setlength{\textwidthB}{0.83\textwidth}

\def\VideoName{moorea\_2019-02-19\_005\_td\_915500000\_975500000}
\def\Img#1#2{figures/\VideoName/#1/\VideoName\_#2.pdf}
\def\VideoNameD{moorea\_2019-02-18\_000\_td\_2867500000\_2927500000}
\def\ImgD#1#2{figures/\VideoNameD/#1/\VideoNameD\_#2.pdf}
\def\VideoNameF{moorea\_2019-02-18\_000\_td\_2806500000\_2866500000}
\def\ImgF#1#2{figures/\VideoNameF/#1/\VideoNameF\_#2.pdf}


\begin{figure}[tb]
  \centering
  \makebox[\textwidthA][l]{\raisebox{1.2\baselineskip}{\strut Ground Truth}}
  \begin{subfigure}[b]{\textwidthA}
    \includegraphics[width=\linewidth]{\Img{gt}{404}}
  \end{subfigure}
  \begin{subfigure}[b]{\textwidthA}
    \includegraphics[width=\linewidth]{\Img{gt}{445}}
  \end{subfigure}
  \begin{subfigure}[b]{\textwidthA}
    \includegraphics[width=\linewidth]{\Img{gt}{456}}
  \end{subfigure}
  \begin{subfigure}[b]{\textwidthA}
    \includegraphics[width=\linewidth]{\Img{gt}{1098}}
  \end{subfigure}
  \begin{subfigure}[b]{\textwidthA}
    \includegraphics[width=\linewidth]{\Img{gt}{1128}}
  \end{subfigure}

  \makebox[\textwidthA][l]{\raisebox{1.2\baselineskip}{\strut RVT\cite{RVT}}}
  \begin{subfigure}[b]{\textwidthA}
    \includegraphics[width=\linewidth]{\Img{RVT}{404}}
  \end{subfigure}
  \begin{subfigure}[b]{\textwidthA}
    \includegraphics[width=\linewidth]{\Img{RVT}{446}}
  \end{subfigure}
  \begin{subfigure}[b]{\textwidthA}
    \includegraphics[width=\linewidth]{\Img{RVT}{456}}
  \end{subfigure}
  \begin{subfigure}[b]{\textwidthA}
    \includegraphics[width=\linewidth]{\Img{RVT}{1099}}
  \end{subfigure}
  \begin{subfigure}[b]{\textwidthA}
    \includegraphics[width=\linewidth]{\Img{RVT}{1129}}
  \end{subfigure}

  \makebox[\textwidthA][l]{\raisebox{2.1\baselineskip}{\strut\parbox{\textwidthA}{Prior Works \\ \cite{RED, DMANet, HMNet}}}}
  \begin{subfigure}[b]{\textwidthA}
    \includegraphics[width=\linewidth]{\Img{RED_model_31}{404}}
    \caption{20.20 s}
    \label{fig:probdef1}
  \end{subfigure}
  \begin{subfigure}[b]{\textwidthA}
    \includegraphics[width=\linewidth]{\Img{RED_model_31}{445}}
    \caption{22.25 s}
    \label{fig:probdef3}
  \end{subfigure}
  \begin{subfigure}[b]{\textwidthA}
    \includegraphics[width=\linewidth]{\Img{RED_model_31}{456}}
    \caption{22.80 s}
    \label{fig:probdef4}
  \end{subfigure}
  \begin{subfigure}[b]{\textwidthA}
    \includegraphics[width=\linewidth]{\Img{RED_model_31}{1098}}
    \caption{54.90 s}
    \label{fig:probdef5}
  \end{subfigure}
  \begin{subfigure}[b]{\textwidthA}
    \includegraphics[width=\linewidth]{\Img{RED_model_31}{1128}}
    \caption{56.40 s}
    \label{fig:probdef6}
  \end{subfigure}



  \vspace{0.3cm}
  \makebox[\textwidthA][l]{\raisebox{2.2\baselineskip}{\strut Prior Works}}
  \begin{subfigure}[b]{\textwidthB}
    \includegraphics[width=\linewidth]{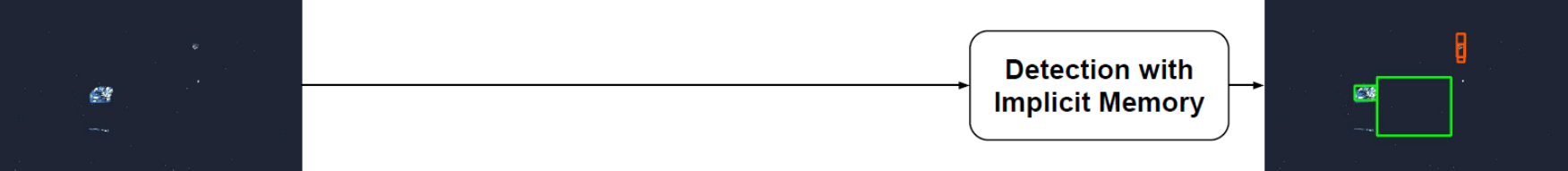}
    \caption{Functional blocks of prior works.}
    \label{fig:priorwork}
  \end{subfigure}

  \vspace{0.3cm}
  \makebox[\textwidthA][l]{\raisebox{5.2\baselineskip}{\strut TEDNet}}
  \begin{subfigure}[b]{\textwidthB}
    \includegraphics[width=\linewidth]{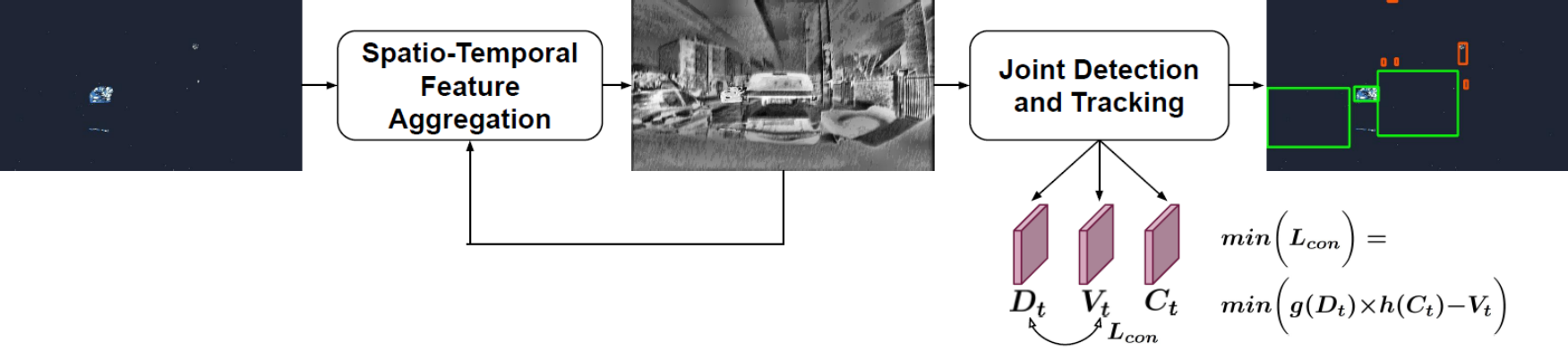}
    \caption{Functional blocks of our TEDNet.}
    \label{fig:tednet}
  \end{subfigure}

  \caption{A video sequence from the 1 Megapixel Automotive Detection Dataset with absolute timestamps, ground truth labels, and some detection results from prior works. Both cars slow down from \ref{fig:probdef1} to \ref{fig:probdef3}, remain still from \ref{fig:probdef4} to \ref{fig:probdef5} and start moving at \ref{fig:probdef6}. Meanwhile, \ref{fig:priorwork} and \ref{fig:tednet} show the main difference between prior works and our TEDNet. Bounding boxes with different colors correspond to different categories.}
  \label{fig:probdef}
\end{figure}





Prior works propose different kinds of memory architectures to retain object permanence in event-based object detection\cite{RED, RVT, ASTMNet, DMANet, HMNet}. 
One common characteristic among these models is that they all use implicit memories with different spatial or temporal resolutions. After some experiments\footnotemark{} shown in \cref{fig:probdef},
\footnotetext{We cannot experiment on ASTMNet\cite{ASTMNet} due to no available source code. HMNet\cite{HMNet} is trained on GEN1 Automotive Detection Dataset with a smaller spatial resolution (304 x 240). Therefore, we downsample the spatial resolution of 1 Megapixel Automotive Detection Dataset (1280 x 720) and test them on HMNet\cite{HMNet} directly.}
we found that only RVT\cite{RVT} can better handle the long-range dependency problem, i.e., objects become invisible when there are no relative motions between objects and the camera.
Meanwhile, RVT\cite{RVT} produces many false positives with high confidence scores due to its implicit ConvLSTMs. Other models in \cref{fig:probdef} are memoryless as long as the duration of missing features is too long, which is over 30 seconds from \cref{fig:probdef4} to \cref{fig:probdef5} and degrades the mAP performance of object detection.

In this work, we propose to formulate the long-range dependency problem for still objects in event cameras as an occlusion problem and use tracking through occlusion\cite{TrackingOcclusions2005, TrackingOcclusions2010_1, TrackingOcclusions2010_2, PermaTrack, RAM} to keep the object permanence. Intuitively, some analogies exist between the long-range dependency problem for still objects and the tracking through occlusion. First, there are no features of the objects we want to detect in both cases. Second, tracking through occlusion is based on a constant velocity assumption; instead, the long-range dependency problem for still objects is based on a zero velocity assumption. Hence, tracking through occlusion can potentially solve the long-range dependency problem for still objects in event cameras. To our knowledge, we are the first to regard the long-range dependency problem in event cameras as a tracking-through-occlusion paradigm.

In order to supervise the training of object permanence, we need additional \emph{visibility} labels to distinguish between moving objects and still objects. Hence, we propose an algorithm to perform the annotation automatically. Furthermore, our model is based on some joint object detection and tracking models, which give object detectors tracking ability. We augment these models with a spatio-temporal feature aggregation module and a consistency loss to not only enrich the latent features as shown in \cref{fig:tednet} but also increase the robustness of the overall pipeline where still objects are retained but real occluded objects are discarded. In summary, our model outperforms state-of-the-art event-based object detectors by 7.9\% absolute mAP, and our main contributions are as follows:

\begin{enumerate}
\item We propose an innovative insight into memorizing missing features in event cameras by using tracking as an explicit memory instead of the traditional implicit memory method.
\item We propose an auto-labeling algorithm to provide a more reliable dataset for future works, leading to better training results without costly manual labor.
\item We propose a spatio-temporal feature aggregation module and a consistency loss to track still or pseudo-occluded objects instead of real occluded objects and further improve the mAP performance.
\item The efficacy of the proposed architecture has been experimented on 1 Megapixel Automotive Detection Dataset. It significantly surpasses state-of-the-art event-based object detectors by 7.9\% absolute mAP.
\end{enumerate}

\section{Related Works}

\subsection{Event Representations}

Event representations can be divided into sparse and dense representations according to the downstream tasks. For classification tasks, sparse or point-cloud-like representations are more commonly used; instead, for localization tasks, dense or image-like representations are more widely used except for a small portion of works that apply graphs\cite{AEGNN} or spikes\cite{SNN2022, SNN2023} as an event representation for object detection.

Dense representations can be further divided into handcrafted methods and learning-based methods. There are four mainstream handcrafted methods, including \emph{Event Histogram}\cite{EventHistogram2018}, \emph{Timestamp}\cite{Timestamp2016, Timestamp2018, Timestamp2020, Timestamp2022}, \emph{Time Surface}\cite{TimeSurface2016, TimeSurface2018, TimeSurface2020}, and \emph{Event Volume}\cite{EventVolume2018, EventVolume2019_1, EventVolume2019_2, EventVolume2020}. RED\cite{RED} has validated that \emph{Event Volume} gets better mAP performance in event-based object detection compared with other handcrafted event representations.

For learning-based methods, EST\cite{EST} and Matrix-LSTM\cite{MatrixLSTM} are designed for classification tasks instead of localization tasks. In addition, ASTMNet\cite{ASTMNet}, DMANet\cite{DMANet}, and HMNet\cite{ASTMNet} propose different learning-based event representations for object detection and get some mAP performance improvements.

\subsection{Object Detection with Event Cameras}
Object detection with event cameras can be categorized into three branches according to the event representations and network architectures, including GNN-based, SNN-based, and DNN-based methods. GNN-based\cite{AEGNN} and SNN-based\cite{SNN2022, SNN2023} methods are more challenging due to inferior message propagation through a graph and difficulties in optimization, respectively. DNN-based methods\cite{RED, ASTMNet, RVT, DMANet, HMNet} convert sparse events in a time window to a dense event representation that discards some temporal information due to sampling and apply deep neural networks for object detection. Hence, how to obtain better spatio-temporal features across adjacent frames becomes crucial to enhance the performance of event-based object detection, which is very similar to video object detection tasks\cite{VOD2019, VOD2020, VOD2022_1, VOD2022_2}. Different kinds of memory architectures are proposed to retain better spatio-temporal features, including multi-stage ConvLSTMs in RED\cite{RED} and RVT\cite{RVT}, TACN and a single-stage ConvLSTM in ASTMNet\cite{ASTMNet}, long and short memories in DMANet\cite{DMANet}, and an attention-based hierarchical memory in HMNet\cite{HMNet}. Although these memory architectures can memorize historical information for a while, most of them are still forgettable after a very long time and degrade the mAP performance of object detections.

\subsection{Multi-Object Tracking}

Multi-object tracking can be divided into two branches, including \emph{tracking by detection} and \emph{joint object detection and tracking}, where object detection and bounding box association are performed separately and jointly, respectively. The most challenging problem in multi-object tracking is \emph{tracking with long-term occlusions} where tracking identity needs to be maintained even though occlusion incurs temporal invisibility.

\subsubsection{Tracking by Detection} is a two-stage method where object detection and bounding box association are performed separately. It can be further categorized into online\cite{TBDon2009, SORT, DeepSORT, TBDon2019} and offline\cite{TBDoff2006, TBDoff2011_1, TBDoff2011_2, TBDoff2014, TBDoff2020} approaches according to whether the current bounding box association can refer to the future bounding boxes or not. In this work, we focus on online approaches because they are causal and more practical in real scenarios. SORT\cite{SORT} and DeepSORT\cite{DeepSORT} are two classical works to apply the Kalman filter for bounding box association. The only difference between these two works is that the bounding box association policy in SORT\cite{SORT} is based on the intersection over union (IOU) of two bounding boxes, and the association policy in DeepSORT\cite{DeepSORT} is based on the appearance features from a deep neural network, which makes it more robust to deal with occlusion in longer periods of time. These rule-based approaches lack generalizability and make them harder to transfer to different scenarios.


\subsubsection{Joint Object Detection and Tracking} is a one-stage and online method where object detection and bounding box association are performed jointly by an end-to-end trainable deep neural network. \cite{JDT2019} converts Faster R-CNN\cite{FasterRCNN} to a \emph{Tracktor} by utilizing bounding box regression to predict the bounding boxes of the next frame. CenterTrack\cite{CenterTrack} is built on top of CenterNet\cite{CenterNet_arXiv, CenterNet_CVPR} and outputs not only bounding box locations but also tracking vectors, i.e., a vector from the bounding box center of the current frame to the corresponding bounding box center of the previous frame. PermaTrack\cite{PermaTrack} is built on top of CenterTrack\cite{CenterTrack} and is able to track invisible objects for a long-term feature disappearance by supervising the visibility property during the training stage. RetinaTrack\cite{RetinaTrack} is built on top of RetinaNet\cite{RetinaNet} and is able to track occluded objects by using instance-level features to perform bounding box association. These learning-based approaches can better keep the object permanence in the presence of long-term occlusions.


\section{Method}
\begin{figure}[tb] 
  \centering
  \includegraphics[width=0.9\linewidth]{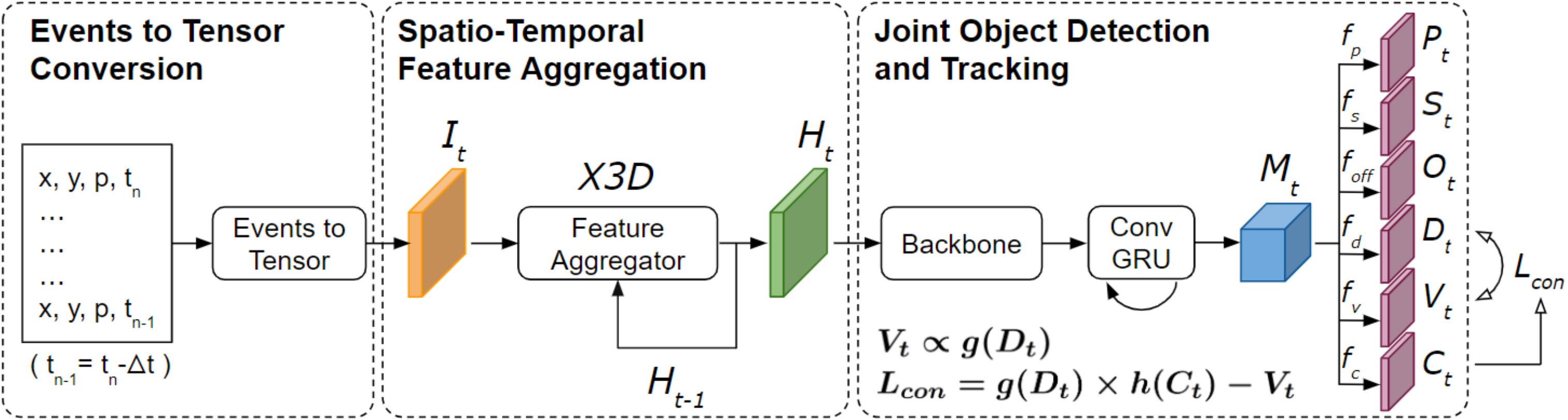}
  \caption{The architecture of the proposed TEDNet. Spatio-Temporal Feature Aggregation integrates a 3D convolution with a recurrent neural network where $H_t = \emph{X3D}(I_t,\ H_{t-1})$. Joint Object Detection and Tracking consists of localization head $f_p$/map $P_t$, size head $f_s$/map $S_t$, offset head $f_{off}$/map $O_t$, displacement head $f_d$/map $D_t$, visibility head $f_v$/map $V_t$, and consistency head $f_c$/map $C_t$. The novel consistency map $C_t$ correlates the proportional relationship between $D_t$ and $V_t$ where the novel consistency loss $L_{con}$ offers a better regularization to harness the advantage of X3D and increase the robustness of the overall pipeline.}
  \label{fig:architecture}
\end{figure}

Our \textbf{T}racking-assisted \textbf{E}vent-based object \textbf{D}etector is denoted \textbf{TEDNet} which consists of four modules, including \emph{event-to-tensor conversion}, \emph{spatio-temporal feature aggregation}, \emph{joint object detection and tracking}, and \emph{consistency loss} as shown in \cref{fig:architecture}. First, input events are sampled with a constant time window and converted to a dense event representation. Then, the dense tensor is passed through a \emph{spatio-temporal feature aggregator} to retain as much spatio-temporal information as possible, especially for those still objects with no features. After that, a \emph{joint object detection and tracking} module is used to track those still objects according to the object permanence property. Finally, a \emph{consistency loss} correlates two feature maps and increases the robustness of the overall pipeline. In addition, an auto-labeling algorithm is proposed to supervise the training of object permanence by distinguishing between moving objects and still objects. All of the details are elaborated in the latter parts of this section.

\subsection{Auto-Labeling for Still Objects}
\label{sec:autolabel}
The labels for object detection in 1 Megapixel Automotive Detection Dataset come from the inference results of its corresponding RGB images. It offers only bounding boxes and tracking identities, which are insufficient for an object detector to keep object permanence for still objects. Therefore, we propose an algorithm to split objects into still objects and moving objects automatically to supervise the training of object permanence, as shown in \cref{alg:autolabel}. In a nutshell, still objects are invisible by an event camera where \emph{visibility} labels are annotated as 0.0. However, moving objects are visible by an event camera where \emph{visibility} labels are annotated as 1.0. The criteria to decide whether an object is still are based on the feature sparsity in the bounding box and the displacement between centers of two bounding boxes with the same tracking identity in adjacent frames. In the following, the overall procedures are divided into three parts, including \emph{calculation of occupancy rate}, \emph{calculation of displacement}, and \emph{continuity maintenance}.

\subsubsection{Calculation of Occupancy Rate} is to determine whether the features of an object are sparse or not. First, input features of shape $[C,\ H,\ W]$ for $i$-th frame are converted to a binary mask dubbed \emph{occupancy\_mask} in \cref{alg:occupancy_mask} of \cref{alg:autolabel}. The \emph{occupancy\_mask} is a tensor of shape $[H,\ W]$ indicating whether at least one event appears at a specific pixel location in the time duration $[(i-1)\times \Delta t,\ i\times \Delta t)$ or not, where $i\in[1,\ T]$ is the frame identity and $\Delta t$ is the constant time window. Then, \emph{occ\_mask} is obtained by slicing the corresponding bounding box region in \cref{alg:occ_mask}. Next, \emph{bbox\_mask} is obtained by discarding the overlapping regions with other bounding boxes in \cref{alg:bbox_mask}. In more detail, \emph{bbox\_mask} is a tensor of shape $[h_j,\ w_j]$, which is the same as \emph{occ\_mask}, where all pixel values are initialized to 1 and pixel values in those locations that overlap with other bounding boxes are turned to 0. After that, \emph{occ\_true} is calculated by \cref{eq:occ_true} representing the number of most possible occupied pixels for the corresponding object, and \emph{occ\_false} is calculated by \cref{eq:occ_false} representing the number of most certain pixels for the background. Finally, \emph{occ\_rate} is obtained by \cref{eq:occ_rate} in \cref{alg:occ_rate}, which is a ratio indicating how much percentage of pixel values of 1 in \emph{bbox\_mask} are occupied by \emph{occ\_mask}. In summary, the occupancy rate represents the feature sparsity of a specific object. Its value is high in moving objects due to more features and is low in still objects due to fewer or no features.

\begin{equation}
\label{eq:occ_true}
occ\_true = \sum_{\substack{0<x<w_j \\0<y<h_j}} (occ\_mask(x, y)==1)\times bbox\_mask(x, y)
\end{equation}

\begin{equation}
\label{eq:occ_false}
occ\_false = \sum_{\substack{0<x<w_j \\0<y<h_j}} (occ\_mask(x, y)==0)\times bbox\_mask(x, y)
\end{equation}

\begin{equation}
\label{eq:occ_rate}
occ\_rate = \frac{occ\_true}{occ\_true+occ\_false}
\end{equation}

\subsubsection{Calculation of Displacement} is to approximate the motion property of the corresponding object by the bounding box center in the current frame $[c_x,\ c_y]$ and that with the same tracking identity in the previous frame $[pc_x,\ pc_y]$ from \cref{alg:vel_start} to \ref{alg:vel_end}. The displacement is normalized by the width and height of the current bounding box by \cref{eq:vel} in \cref{alg:vel_end} to discard the drifting effect of inaccurate bounding boxes for both small and large still objects. As a matter of fact, this approximation is inaccurate due to the lack of depth information in event cameras. Therefore, the decision-making of still objects depends on not only displacement but also occupancy rate.

\begin{equation}
\label{eq:vel}
disp = \sqrt{(\frac{pc_x-c_x}{w_j})^2+(\frac{pc_y-c_y}{h_j})^2}
\end{equation}

\subsubsection{Continuity Maintenance} is used to make the variation of \emph{visibility} label for each specific tracking identity as smooth as possible and discard those bounding boxes with no features at their first occurrence. If there is no corresponding bounding box with the same tracking identity in the previous frame, a \emph{visibility} label is determined by only the occupancy rate from \cref{alg:first_occur_start} to \ref{alg:first_occur_end}. Also, a buffer dubbed \emph{still\_hits} consists of several key-value pairs where keys are tracking identities, and values are initialized to 0 and limited to 5, representing the number of still hits for each tracking identity. \emph{still\_hits} can help to smooth out the impulse of \emph{visibility} variation owing to the original label noises. If there is a corresponding tracked object in the previous frame, a \emph{visibility} label is determined by either both occupancy rate and displacement from \cref{alg:prev_track_start} to \ref{alg:prev_track_mid1} or only the \emph{still\_hits} buffer from \cref{alg:prev_track_mid2} to \ref{alg:prev_track_end}. Finally, not only visible objects but also those invisible objects that are tracked for two previous frames are added to the new labels from \cref{alg:track_vis_start} to \ref{alg:track_vis_end} where \emph{track\_vis} consists of several key-value pairs where keys are tracking identities, and values are queues with size 2, representing the continuity of frame identities. As a result, only $N_i$ out of $M_i$ objects are preserved with additional \emph{visibility} labels for $i$-th frame.

\SetKwComment{Comment}{/* }{ */}

\newlength\mylen
\newcommand\myinput[1]{%
  \settowidth\mylen{\KwIn{}}%
  \setlength\hangindent{\mylen}%
  \hspace*{\mylen}#1\\}

\let\oldnl\nl
\newcommand{\nonl}{\renewcommand{\nl}{\let\nl\oldnl}}

\begin{algorithm}
\caption{Auto-labeling for still objects}
\label{alg:autolabel}
\SetAlgoLined

\KwIn{$inputFeatures$ = tensor of shape $[T,\ C,\ H,\ W]\ ;\ i\in[1,\ T]$}
\nonl\myinput{$oldLabels[i] = \{[x_j,\ y_j,\ w_j,\ h_j,\ trackId_j]\ |\ j\in[1,\ M_i]\}$}

\KwOut{$newLabels[i] = \{[x_k,\ y_k,\ w_k,\ h_k,\ trackId_k,\ visibility_k]\ |\ k\in[1,\ N_i]\}$}

 $\Delta t \gets 50000\ \mu s;\ still\_hits \gets \{\};\ track\_vis \gets \{\}$\;
 \For{$i\gets1$ \KwTo T}{
  $newLabels[i] \gets \{\}$\;
  $occupancy\_mask$ = get\_occupancy\_mask($inputFeatures[i]$)\; \label{alg:occupancy_mask}
  \ForEach{$[x_j,\ y_j,\ w_j,\ h_j,\ trackId_j] \in oldLabels[i]$}{
   $occ\_mask = occupancy\_mask[x_j:x_j+w_j,\ y_j:y_j+h_j]$\; \label{alg:occ_mask}
   $bbox\_mask$ = get\_bbox\_mask($x_j,\ y_j,\ w_j,\ h_j,\ oldLabels[i]$)\; \label{alg:bbox_mask}
   $occ\_rate$ = get\_occupancy\_rate($occ\_mask,\ bbox\_mask$)\; \label{alg:occ_rate}
   $visibility_j \gets 1.0$ \Comment*[r]{Initialize to a moving object.}
   \uIf{$trackId_j$ \textbf{in} $newLabels[i-1]$}{
    $c_x,\ c_y$ = get\_bbox\_center($x_j,\ y_j,\ w_j,\ h_j$)\; \label{alg:vel_start}
    $x_k,\ y_k,\ w_k,\ h_k,\ trackId_k,\ visibility_k$ = get\_previous\_bbox($trackId_j,\ newLabels[i-1]$)\;
    $pc_x,\ pc_y$ = get\_bbox\_center($x_k,\ y_k,\ w_k,\ h_k$)\;
    $disp$ = get\_normalized\_displacement($pc_x,\ pc_y,\ c_x,\ c_y$)\; \label{alg:vel_end}
    \uIf{$disp < D_{value}$ \textbf{and} $occ\_rate < O_{value}$}{ \label{alg:prev_track_start}
     $visibility_j \gets 0.0$ \Comment*[r]{Change to a still object.}
     $still\_hits[trackId_j]\gets still\_hits[trackId_j]+1$\;
    } \label{alg:prev_track_mid1}
    \ElseIf{$still\_hits[trackId_j] > 0$}{ \label{alg:prev_track_mid2}
     $visibility_j \gets 0.0$ \Comment*[r]{Change to a still object.}
     $still\_hits[trackId_j]\gets still\_hits[trackId_j]-1$\; \label{alg:prev_track_end}
    }
   }
   \ElseIf{$occ\_rate < O_{value}$}{ \label{alg:first_occur_start}
    $visibility_j \gets 0.0$ \Comment*[r]{Change to a still object.}
    $still\_hits[trackId_j]\gets still\_hits[trackId_j]+1$\; \label{alg:first_occur_end}
   }
   \If{$visibility_j = 1.0$ \textbf{or} $track\_vis[trackId_j]=[i-2,\ i-1]$}{ \label{alg:track_vis_start}
    $newLabels[i]$.add($[x_j,\ y_j,\ w_j,\ h_j,\ trackId_j,\ visibility_j]$)\;
    $track\_vis[trackId_j]$.enqueue($i$)\; \label{alg:track_vis_end}
   }
  }
 }
\end{algorithm}

\subsection{Event Representation}
Prior works have proved that learning-based event representations\cite{ASTMNet, DMANet, HMNet} outperform handcrafted event representations\cite{RED, RVT}, but learning-based methods have higher computational complexity, which makes training slow. In this work, we do not focus on designing new event representations; instead, we opt for the best-performing handcrafted event representation\cite{RED}, i.e., \emph{Event Volume}, as our \emph{event-to-tensor conversion} module. The benefit of using handcrafted methods is that they can be preprocessed in advance instead of on-the-fly processing like learning-based methods. As a result, we can focus on retaining object permanence to enhance object detection performance with faster training speed.

\subsection{Spatio-Temporal Feature Aggregation}
The most significant challenge in event cameras is the long-range dependency problem. In this work, we propose to apply either a simple 3D convolutional network(\emph{C3D})\cite{C3D} or a deformable 3D convolutional network(\emph{D3D})\cite{D3D} as well as a ConvGRU in \emph{joint object detection and tracking} module as our spatio-temporal feature aggregator. Unlike \cite{C3D, D3D} that input a video sequence to get a better spatio-temporal feature for the downstream task, our approach inputs only two frames, i.e., a current frame $I_t$ and a previous history frame $H_{t-1}$. The current history frame is obtained by $H_t = \emph{X3D}(I_t,\ H_{t-1})$ where \emph{X3D} is either \emph{C3D} or \emph{D3D}. To be more specific, we integrate an \emph{X3D} with a recurrent neural network into our novel \emph{recurrent X3D} to get a better spatio-temporal feature as an input to the following \emph{joint object detection and tracking} module.

While both 3D convolution and ConvGRU are effective for aggregating spatio-temporal features, 3D convolution tends to be more computationally efficient, especially when the input feature has high spatial resolution. Therefore, we opt for 3D convolution on the input side where $I_t$ and $H_t$ have the same spatial resolution and ConvGRU on the output side where $M_t$ is downsampled by 4 times from $I_t$ as shown in \cref{fig:architecture}.

\subsection{Joint Object Detection and Tracking}
Although \emph{tracking by detection} is a classical tracking paradigm, it has some intrinsic drawbacks due to its two-stage properties where bounding box association fails if occlusion causes incorrect detection.
Hence, in this work, we propose to apply a one-stage \emph{joint object detection and tracking} method to better leverage tracking capability to enhance object detection performance. Two candidates for the \emph{joint object detection and tracking} method are CenterTrack\cite{CenterTrack} and PermaTrack\cite{PermaTrack}. In addition to the localization head $f_p$, size head $f_s$, and offset head $f_{off}$ in CenterNet\cite{CenterNet_arXiv, CenterNet_CVPR}, CenterTrack\cite{CenterTrack} adds one extra displacement head $f_d$ for tracking, which represents a tracking vector from the bounding box center of the current frame to the corresponding bounding box center of the previous frame. PermaTrack\cite{PermaTrack} adds one more visibility head $f_v$ to represent the occlusion ratio, i.e., if an object is not occluded, fully occluded, or partially occluded, its visibility is 1.0, 0.0, or between 0.0 and 1.0, respectively. In the situation of event cameras, visibility head $f_v$ is used to represent the mobility ratio, i.e., if an object is still or moving, its visibility is 0.0 or 1.0, respectively. There is no visibility value between 0.0 and 1.0 because it is hard to determine the accurate velocity by using only 2D image coordinates as mentioned in \cref{sec:autolabel}.

\subsection{Consistency Loss}
The displacement head $f_d$ outputs a displacement map $D_t\in \mathbb{R}^{\frac{H}{R}\times \frac{W}{R}\times 2}$ to represent the tracking vectors of every object with downsampling factor $R=4$. However, the 
visibility head $f_v$ outputs a visibility map $V_t\in [0,\ 1]^{\frac{H}{R}\times \frac{W}{R}}$ to represent whether the center coordinates $p_i$ of $i$-th detected object is still or not. Both $D_t$ and $V_t$ imply the mobility of objects in two different forms with different channels and magnitudes. To be more concrete, $D_t$ and $V_t$ have some proportional relationship after some transformation, i.e. $V_t \propto g(D_t)$ where $g(D_t)={\|D_t\|}_2\in \mathbb{R}^{\frac{H}{R}\times \frac{W}{R}}$ is an Euclidean norm of $D_t$ representing the magnitude of movements. In this work, we propose another consistency head $f_c$ that outputs a consistency map $C_t$ to model the proportional relationship where the learnable proportional ratio is $h(C_t)=e^{-relu(C_t)}\in [0,\ 1]^{\frac{H}{R}\times \frac{W}{R}}$ and our learning goal is to minimize the difference between $g(D_t) \times h(C_t)$ and $V_t$, i.e. consistency loss, by regression as shown in \cref{eq:proprelation}. As a result, the consistency loss between $D_t$ and $V_t$ is shown in \cref{eq:consloss2} where $p_i$ belongs to visible objects. For invisible objects, both $D^{p_i}_t$ and $V^{p_i}_t$ are very close to 0 where the predicted $C^{p_i}_t$ in \cref{eq:consloss2} can be any value that makes training unstable.


\begin{equation}
\label{eq:proprelation}
min\Big(L_{con}\Big) = min\bigg(\Big|g(D_t) \times h(C_t) - V_t\Big|\bigg) = min\bigg(\Big|{\|D_t\|}_2 \times e^{-relu(C_t)} - V_t\Big|\bigg)
\end{equation}


\begin{equation}
\label{eq:consloss2}
L_{con} = \frac{1}{N} \sum_{i=1}^{N} \bigg|{\Big\|D^{p_i}_t\Big\|}_2\times e^{-relu(C^{p_i}_t)}-V^{p_i}_t\bigg|
\end{equation}

\section{Experiments}

\subsection{Dataset with Auto-Labeling}
Prophesee, a company developing event-based sensors and algorithms, released the first large-scale and high-resolution (1280 $\times$ 720) event-based object detection dataset and its deep neural network RED\cite{RED} for automotive scenarios. This 1 Megapixel Automotive Detection Dataset is captured by the 1-megapixel event camera\cite{evsensor} and a standard RGB camera, but only event data and its annotations are released. The annotations come from the inference result of the RGB image and transfer to the event camera coordinate by geometric transformation. After visualizing the annotations in the dataset, we found that there are several incorrect bounding boxes with either imprecise locations and sizes or incorrect labels. Most importantly, several bounding boxes contain no features at the very beginning of some videos because they are still and visible by an RGB camera but invisible by an event camera, and humans cannot recognize their physical existence with only event data even though we have temporal cues. These incorrect bounding boxes limit the object detection performance and make model training awkward. Therefore, we clean the dataset by an auto-labeling algorithm in \cref{sec:autolabel} with hyperparameters $D_{value}$=0.03 and $O_{value}$=0.1 to not only insert additional \emph{visibility} labels to supervise the training of object permanence but also discard those bounding boxes that are impossible to be detected by humans. We named the dataset before/after data cleaning \emph{noisy GT}/\emph{clean GT} and its corresponding model \emph{noisy model}/\emph{clean model}.


To validate the effectiveness of the proposed auto-labeling algorithm, we train and test the RED\cite{RED} on both \emph{noisy GT} and \emph{clean GT} by using the training and testing set respectively, and evaluate the performance by using mAP@0.5 as shown in \cref{tab:datacleaning}. The first row shows that data cleaning does not degrade the model performances on \emph{noisy GT}. The first or second column shows that mAP increases on \emph{clean GT} because those impossible-to-detect objects are discarded successfully. The second row shows that mAP increases with \emph{clean model} because the auto-labeling algorithm effectively makes model training more robust.

\begin{table}[tb]
  \caption{The effectiveness of the auto-labeling algorithm by RED\cite{RED}}
  \label{tab:datacleaning}
  \centering
  \begin{tabular}{P{1.5cm} P{1.5cm} P{1.5cm} P{1.5cm}}
    \toprule
    \multicolumn{2}{c}{\multirow{2}{*}{mAP@0.5}} & \multicolumn{2}{c}{Train on} \\
    \multicolumn{2}{c}{} & noisy GT & clean GT \\
    \midrule
    \multirow{2}{*}{Test on} & noisy GT & 43.5 & 43.6 \\
    & clean GT & 47.1 & \textbf{48.2} \\
    \bottomrule
  \end{tabular}
\end{table}

\subsection{Ablation Study}
The models in ablation studies are trained on \emph{clean GT} and tested on both \emph{noisy GT} and \emph{clean GT} as shown in \cref{tab:ablation}. We choose two \emph{joint object detection and tracking} models as our baselines, CenterTrack and PermaTrack\cite{PermaTrack}, respectively.

\begin{table}[tb]
  \caption{Ablation study of each component in our TEDNet}
  \label{tab:ablation}
  \centering
  \begin{tabular}{p{1.7cm} P{1cm} P{1cm} P{1cm} P{1cm} P{1cm} P{1cm} P{1cm} P{1cm} P{1.5cm}}
    \toprule
    Methods & (a) & (b) & (c) & (d) & (e) & (f) & (g) & (h) & \textbf{TEDNet} \\ 
    \midrule
    CenterTrack\textsuperscript{*} & \checkmark & \checkmark & \checkmark & & & & & & \\
    PermaTrack & & & & \checkmark & \checkmark & \checkmark & \checkmark & \checkmark & \checkmark \\
    D3D & & \checkmark & & & \checkmark & & & & \checkmark \\
    C3D & & & \checkmark & & & \checkmark & & \checkmark &\\
    $L_{con}$ & & & & & & & \checkmark & \checkmark & \checkmark \\
    \midrule
    mAP@0.5 & & & & & & & & \\
    noisy GT & 48.3 & 53.9 & 53.4 & 48.8 & 53.8 & 55.0 & 51.1 & 54.1 & \textbf{56.2} \\
    clean GT & 54.1 & 59.7 & 59.2 & 55.2 & 60.5 & 61.2 & 57.7 & 60.8 & \textbf{62.8} \\
    \bottomrule
    \multicolumn{10}{l}{* The CenterTrack here is the original one\cite{CenterTrack} with ConvGRU, making it comparable} \\
    \multicolumn{10}{l}{\ \ with PermaTrack\cite{PermaTrack}.}
  \end{tabular}
\end{table}

\subsubsection{Spatio-Temporal Feature Aggregation}
We study two \emph{spatio-temporal feature aggregation} approaches along with CenterTrack\cite{CenterTrack} and PermaTrack\cite{PermaTrack}. Comparing (b) with (c), \emph{D3D} performs slightly better than \emph{C3D} if we use CenterTrack\cite{CenterTrack} as the baseline. However, comparing (e) with (f), we are surprised that \emph{C3D} is slightly superior to \emph{D3D} if we use PermaTrack\cite{PermaTrack} as the baseline. Intuitively, \emph{C3D} is the performance lower bound of \emph{D3D} with the same convolution kernel size (3$\times$3), but \emph{D3D} introduces additional parameters to learn offsets, making it harder to converge. Therefore, we need more fine-grained hyperparameter tuning or regularization to harness the advantages of deformable convolutions. In summary, \emph{spatio-temporal feature aggregation} improves the mAP@0.5 by 5.0\% to 6.2\% compared with the corresponding baselines.

\subsubsection{Consistency Loss} We conduct consistency loss between displacement map $D_t$ and visibility map $V_t$ to model their proportional relationship. Consistency loss can only be applied to PermaTrack\cite{PermaTrack} because there is no visibility head in CenterTrack\cite{CenterTrack}. Comparing (d) with (g), the consistency loss enhances the mAP@0.5 by 2.3\% on \emph{noisy GT} and 2.5\% on \emph{clean GT}. Comparing (f) with (h), the consistency loss degrades the mAP@0.5 by 0.9\% on \emph{noisy GT} and 0.4\% on \emph{clean GT}  if we use \emph{C3D} as the \emph{spatio-temporal feature aggregation} module. In contrast, comparing (e) with our TEDNet, the consistency loss improves the mAP@0.5 by 2.4\% on \emph{noisy GT} and 2.3\% on \emph{clean GT} if we use \emph{D3D} as the \emph{spatio-temporal feature aggregation} module. To sum up, consistency loss offers a better regularization to harness the advantages of deformable convolutions and increases the robustness of the overall pipeline.


\subsection{Comparison with the State-of-the-art}
We compare our TEDNet to the state-of-the-art event-based object detectors on both \emph{noisy GT} in \cref{tab:sotaold} and \emph{clean GT} in \cref{tab:sotanew}. Our method outperforms the state-of-the-art model by 7.9\% absolute mAP@0.5 and achieves 56.2\% on \emph{noisy GT} as shown in \cref{tab:sotaold}. In addition, our method outperforms RED\cite{RED}, CenterTrack\cite{CenterTrack}, and PermaTrack\cite{PermaTrack} by 14.6\%, 8.7\%, and 7.6\% absolute mAP@0.5 respectively and achieves 62.8\% on \emph{clean GT} as shown in \cref{tab:sotanew}.

\begin{table}[tb]
  \caption{Performance comparison with the state-of-the-art event-based object detectors on \textbf{noisy GT}. Runtime is measured by GeForce RTX 3090 GPU where ASTMNet\cite{ASTMNet} does not offer source code, but it should be 5.5 times slower than RVT\cite{RVT} and 1.8 times slower than RED\cite{RED} according to \cite{RVT}.}
  \label{tab:sotaold}
  \centering
  \begin{tabular}{p{2.8cm} p{1.7cm} p{1.0cm} p{1.8cm} p{1.5cm} p{2cm}} 
    \toprule
    Methods & mAP@0.5 & mAP & params(M) & GFLOPs & runtime(ms) \\ 
    \midrule
    RED\cite{RED} & 43.5 & 22.5 & 24.1 & 31.2 & 11.5 \\ 
    ASTMNet\cite{ASTMNet} & 48.3 & - & >100\cite{RVT} & - & >20.0 \\ 
    RVT\cite{RVT} & 47.0 & - & \textbf{18.5} & \textbf{15.2} & \textbf{5.6} \\ 
    DMANet\cite{DMANet} & 46.3 & 24.7 & 28.2 & 62.2 & 15.1 \\ 
    CenterTrack\cite{CenterTrack}\textsuperscript{*}\textsuperscript{\dag} & 48.3 & 25.3 & 21.0 & 46.5 & 12.2 \\
    PermaTrack\cite{PermaTrack}\textsuperscript{\dag}  & 48.8 & 26.0 & 21.2 & 46.5 & 12.4 \\
    \textbf{TEDNet(Ours)}\textsuperscript{\dag}  & \textbf{56.2} & \textbf{31.2} & 21.3 & 48.8 & 13.5 \\
    \bottomrule
    \multicolumn{5}{l}{* The CenterTrack here is the original one with ConvGRU.} \\
    \multicolumn{5}{l}{\dag\ These models are trained on clean GT but tested on noisy GT.} \\
    \multicolumn{5}{l}{\ \ \ Other models are trained and tested on noisy GT.}
  \end{tabular}
\end{table}


\begin{table}[tb]
  \caption{Performance comparison with the state-of-the-art event-based object detectors on \textbf{clean GT}.}
  \label{tab:sotanew}
  \centering
  \begin{tabular}{p{3.8cm} p{2cm} p{1.5cm}} 
    \toprule
    Methods & mAP@0.5 & mAP \\ 
    \midrule
    RED\cite{RED} & 48.2 & 25.9 \\ 
    CenterTrack\cite{CenterTrack} & 54.1 & 29.3 \\
    PermaTrack\cite{PermaTrack} & 55.2 & 30.2 \\
    \textbf{TEDNet(Ours)} & \textbf{62.8} & \textbf{35.8} \\
    \bottomrule
  \end{tabular}
\end{table}


\subsection{Visualization of Detection Results}
\label{sec:Visualization}
We visualize the detection results in \cref{tab:visualization} by a video with two cars remaining still for over 30 seconds and some other cars passing in front of them in between. For RED\cite{RED}, RVT\cite{RVT}, DMANet\cite{DMANet}, and HMNet\cite{HMNet}, visualizations are built on top of their visualized image converted from event data directly by Metavision SDK of Prophesee because their latent feature maps contain much larger channels and smaller spatial resolutions, making them hard to visualize. For our TEDNet, visualizations are built on top of their latent feature maps $H_t$, which contain fewer channels and the same spatial resolutions as their input feature maps $I_t$. While the values scatter everywhere in latent feature maps $H_t$ due to \emph{D3D} and hence visualization is difficult, we apply an adaptive histogram equalization method called CLAHE\cite{CLAHE} to each channel in $H_t$ individually to enhance the contrast and choose one with the best appearance. For CenterTrack\cite{CenterTrack} and PermaTrack\cite{PermaTrack}, visualizations are built on top of their input feature maps $I_t$, and CLAHE\cite{CLAHE} is also applied to each channel in which one with the best appearance is chosen to be comparable with TEDNet.

Either one or two false negatives occur in the duration between 3rd and 4th frames of RED\cite{RED}, DMANet\cite{DMANet},  and HMNet\cite{HMNet}. To be more specific, the false negatives last 17.9 seconds in RED\cite{RED}, 32.4 seconds in DMANet\cite{DMANet}, and interweavingly in HMNet\cite{HMNet}, representing the severity of the long-range dependency problem. CenterTrack\cite{CenterTrack} and PermaTrack\cite{PermaTrack} solve the false negatives by tracking still objects. However, they incur additional false positives shown in the 4th frame of CenterTrack\cite{CenterTrack} and PermaTrack\cite{PermaTrack}. These false positives occur when a moving car passes in front of a still car from the 3rd frame and becomes fully occluded in the 4th frame where the tracking module regards that object as a still object and keeps tracking it. To be more concrete, CenterTrack\cite{CenterTrack} and PermaTrack\cite{PermaTrack} are unable to distinguish between still and real occluded objects and hence produce many false positives. In contrast, our TEDNet can retain only still objects and discard real occluded objects because \emph{D3D} and \emph{consistency loss} retain the clearest spatial-temporal information among all other models and make invisible objects visible.

\newlength{\textwidthC}
\newlength{\textwidthD}
\setlength{\textwidthC}{0.2\textwidth}
\setlength{\textwidthD}{0.16\textwidth}

\begin{table}[tb]
\centering
\begin{tblr}{
  cell{6}{2} = {c=4}{},
  cell{7}{2} = {c=4}{},
  rows = {rowsep=0pt},
  columns = {colsep=1pt},
}
\makebox[\textwidthC][l]{\raisebox{1.2\baselineskip}{\strut Ground Truth}} &
\begin{subfigure}{\textwidthD}
    \includegraphics[width=\linewidth]{\Img{gt}{404}}
\end{subfigure} &
\begin{subfigure}{\textwidthD}
    \includegraphics[width=\linewidth]{\Img{gt}{472}}
\end{subfigure} & 
\begin{subfigure}{\textwidthD}
    \includegraphics[width=\linewidth]{\Img{gt}{735}}
\end{subfigure} & 
\begin{subfigure}{\textwidthD}
    \includegraphics[width=\linewidth]{\Img{gt}{774}}
\end{subfigure} &
\begin{subfigure}{\textwidthD}
    \includegraphics[width=\linewidth]{\Img{gt}{1128}}
\end{subfigure} \\
\makebox[\textwidthC][l]{\raisebox{1.2\baselineskip}{\strut RVT\cite{RVT}}} &
\begin{subfigure}{\textwidthD}
    \includegraphics[width=\linewidth]{\Img{RVT}{404_2}}
\end{subfigure} &
\begin{subfigure}{\textwidthD}
    \includegraphics[width=\linewidth]{\Img{RVT}{472}}
\end{subfigure} & 
\begin{subfigure}{\textwidthD}
    \includegraphics[width=\linewidth]{\Img{RVT}{736}}
\end{subfigure} & 
\begin{subfigure}{\textwidthD}
    \includegraphics[width=\linewidth]{\Img{RVT}{774}}
\end{subfigure} &
\begin{subfigure}{\textwidthD}
    \includegraphics[width=\linewidth]{\Img{RVT}{1129_2}}
\end{subfigure} \\
\makebox[\textwidthC][l]{\raisebox{1.2\baselineskip}{\strut\parbox{\textwidthC}{Prior Works \\ \cite{RED, DMANet, HMNet}}}} &
\begin{subfigure}{\textwidthD}
    \includegraphics[width=\linewidth]{\Img{RED_model_31}{404}}
\end{subfigure} &
\begin{subfigure}{\textwidthD}
    \includegraphics[width=\linewidth]{\Img{RED_model_31}{472}}
\end{subfigure} & 
\begin{subfigure}{\textwidthD}
    \includegraphics[width=\linewidth]{\Img{RED_model_31}{735}}
\end{subfigure} & 
\begin{subfigure}{\textwidthD}
    \includegraphics[width=\linewidth]{\Img{RED_model_31}{774}}
\end{subfigure} &
\begin{subfigure}{\textwidthD}
    \includegraphics[width=\linewidth]{\Img{RED_model_31}{1128}}
\end{subfigure} \\
\makebox[\textwidthC][l]{\raisebox{1.2\baselineskip}{\strut\parbox{\textwidthC}{CenterTrack\cite{CenterTrack} \\ PermaTrack\cite{PermaTrack}}}} &
\begin{subfigure}{\textwidthD}
    \includegraphics[width=\linewidth]{\Img{PermaTrack_model_54}{404}}
\end{subfigure} &
\begin{subfigure}{\textwidthD}
    \includegraphics[width=\linewidth]{\Img{PermaTrack_model_54}{472}}
\end{subfigure} & 
\begin{subfigure}{\textwidthD}
    \includegraphics[width=\linewidth]{\Img{PermaTrack_model_54}{735}}
\end{subfigure} & 
\begin{subfigure}{\textwidthD}
    \includegraphics[width=\linewidth]{\Img{PermaTrack_model_54}{774}}
\end{subfigure} &
\begin{subfigure}{\textwidthD}
    \includegraphics[width=\linewidth]{\Img{PermaTrack_model_54}{1128}}
\end{subfigure} \\
\makebox[\textwidthC][l]{\raisebox{1.2\baselineskip}{\strut TEDNet(Ours)\textsuperscript{*}}} &
\begin{subfigure}{\textwidthD}
    \includegraphics[width=\linewidth]{\Img{TEDNet_model_96}{404}}
\end{subfigure} &
\begin{subfigure}{\textwidthD}
    \includegraphics[width=\linewidth]{\Img{TEDNet_model_96}{472}}
\end{subfigure} & 
\begin{subfigure}{\textwidthD}
    \includegraphics[width=\linewidth]{\Img{TEDNet_model_96}{735}}
\end{subfigure} & 
\begin{subfigure}{\textwidthD}
    \includegraphics[width=\linewidth]{\Img{TEDNet_model_96}{774}}
\end{subfigure} &
\begin{subfigure}{\textwidthD}
    \includegraphics[width=\linewidth]{\Img{TEDNet_model_96}{1128}}
\end{subfigure} \\
 & $\xrightarrow{\hspace*{2cm}}$ time  &  &  &  &  \\
\end{tblr}

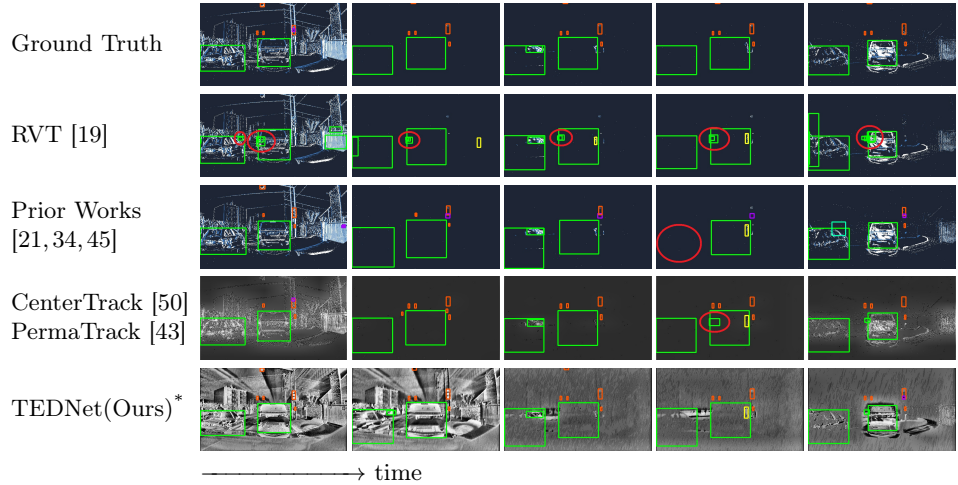
\captionof{figure}{The visualization results of prior works\cite{RED, RVT, DMANet, HMNet, CenterTrack, PermaTrack} and our TEDNet. Bounding boxes with different colors correspond to different categories. Red circles correspond to either false positives or false negatives. TEDNet achieves state-of-the-art mAP performance by retaining bounding boxes of still objects and discarding bounding boxes of real occluded objects.  (Note: * shows the post-processed feature map after the spatio-temporal feature aggregation.)}
\label{tab:visualization}
\end{table}

\section{Conclusion}
We introduce a novel architecture to apply tracking through occlusion to solve the long-range dependency problem for still objects in event cameras. The architecture retains the most abundant spatial-temporal information and tracks only still or pseudo-occluded objects instead of real occluded objects. Our comprehensive experiments demonstrate the effectiveness of each component of our TEDNet that achieves state-of-the-art performance in event-based object detection. In addition, the proposed auto-labeling algorithm provides a more reliable dataset and makes model training more robust.

\section*{Acknowledgements}
This work was supported in part by National Science and Technology Council, Taiwan, under Grant NSTC 111-2634-F-002-022 and by Qualcomm through a Taiwan University Research Collaboration Project.


%
%
\newpage
\bibliographystyle{splncs04}
\bibliography{main}

\newpage

\section*{\centering\large Tracking-Assisted Object Detection with Event Cameras}
\section*{\centering\large Supplementary Material}

\vspace*{1\baselineskip}

In the supplementary material, we provide dataset details, implementation details, visualization of failure cases, additional video, and some quantitative analysis to demonstrate why our TEDNet outperforms prior works by such significant amounts.

\setcounter{section}{0} 
\section{Dataset Details}
1 Megapixel Automotive Detection Dataset is the first large-scale and high-resolution (1280 $\times$ 720) event-based object detection dataset released by Prophesee, a company developing event-based sensors. The annotations of this dataset come from the inference result of the corresponding RGB image and are transferred to the event camera coordinate by geometric transformation, where only event data and its transformed annotations are released. The annotations consist of more than 25 million bounding boxes of cars, pedestrians, and two-wheelers recorded in 60Hz. In addition, bounding boxes of trucks, buses, traffic signs, and traffic lights are also provided. The dataset can be split into training, validation, and testing, with 11.19, 2.21, and 2.25 hours of recording, respectively. Each recording is split into 60-second chunks to increase the training and evaluation efficiency. To speed up the training, each annotation is preprocessed to record not only the bounding boxes at the current timestamp but also the corresponding bounding boxes at the previous timestamp as the ground truth of the tracking vector.

As mentioned in \cref{sec:autolabel}, an auto-labeling algorithm is proposed to not only clean the impossible-to-detect objects as shown in \cref{tab:visualization_noisy_vs_clean_gt} but also add additional labels to distinguish between still and moving objects for the supervision of training object permanence as shown in \cref{tab:visualization_visibility}.

\begin{table}[!hbt]
\centering
\begin{tblr}{
  row{5} = {font=\footnotesize, c}, 
  cell{1}{1} = {c=5}{},
  cell{3}{1} = {c=5}{},
  rows = {rowsep=0pt},
  columns = {colsep=1pt},
}
\centering
Noisy Ground Truth (Noisy GT) &  &  &  &  &  \\
\begin{subfigure}{\textwidthC}
    \includegraphics[width=\linewidth]{\ImgD{noisy_gt}{10}}
\end{subfigure} &
\begin{subfigure}{\textwidthC}
    \includegraphics[width=\linewidth]{\ImgD{noisy_gt}{340}}
\end{subfigure} & 
\begin{subfigure}{\textwidthC}
    \includegraphics[width=\linewidth]{\ImgD{noisy_gt}{650}}
\end{subfigure} & 
\begin{subfigure}{\textwidthC}
    \includegraphics[width=\linewidth]{\ImgD{noisy_gt}{665}}
\end{subfigure} &
\begin{subfigure}{\textwidthC}
    \includegraphics[width=\linewidth]{\ImgD{noisy_gt}{679}}
\end{subfigure} \\
Clean Ground Truth (Clean GT) &  &  &  &  &  \\
\begin{subfigure}{\textwidthC}
    \includegraphics[width=\linewidth]{\ImgD{clean_gt}{10}}
\end{subfigure} &
\begin{subfigure}{\textwidthC}
    \includegraphics[width=\linewidth]{\ImgD{clean_gt}{340}}
\end{subfigure} & 
\begin{subfigure}{\textwidthC}
    \includegraphics[width=\linewidth]{\ImgD{clean_gt}{650}}
\end{subfigure} & 
\begin{subfigure}{\textwidthC}
    \includegraphics[width=\linewidth]{\ImgD{clean_gt}{665}}
\end{subfigure} &
\begin{subfigure}{\textwidthC}
    \includegraphics[width=\linewidth]{\ImgD{clean_gt}{679}}
\end{subfigure} \\
(a) 0.50 s & (b) 17.00 s & (c) 32.50 s & (d) 33.25 s & (e) 33.95 s \\
$\xrightarrow{\hspace*{2.2cm}}$ & time \\
\end{tblr}
\captionof{figure}{A video sequence of the noisy and clean ground truth with absolute timestamps and ground truth labels. One car on the left remains still from the start of the video with no features, and human beings are unaware of the physical existence of this car. Hence, the bounding box of that car is removed according to the auto-labelling algorithm mentioned in \cref{sec:autolabel}. Bounding boxes with different colors correspond to different categories.}
\label{tab:visualization_noisy_vs_clean_gt}
\end{table}

\begin{table}[!hbt]
\centering
\begin{tblr}{
  row{3} = {font=\footnotesize, c}, 
  cell{1}{1} = {c=5}{},
  rows = {rowsep=0pt},
  columns = {colsep=1pt},
}
\centering
Clean Ground Truth with Visibility Labels &  &  &  &  &  \\
\begin{subfigure}{\textwidthC}
    \includegraphics[width=\linewidth]{\Img{gt_visibility}{431}}
\end{subfigure} &
\begin{subfigure}{\textwidthC}
    \includegraphics[width=\linewidth]{\Img{gt_visibility}{436}}
\end{subfigure} & 
\begin{subfigure}{\textwidthC}
    \includegraphics[width=\linewidth]{\Img{gt_visibility}{773}}
\end{subfigure} & 
\begin{subfigure}{\textwidthC}
    \includegraphics[width=\linewidth]{\Img{gt_visibility}{1110}}
\end{subfigure} &
\begin{subfigure}{\textwidthC}
    \includegraphics[width=\linewidth]{\Img{gt_visibility}{1132}}
\end{subfigure} \\
(a) 21.55 s & (b) 21.80 s & (c) 38.65 s & (d) 55.50 s & (e) 56.60 s \\
$\xrightarrow{\hspace*{2.2cm}}$ & time \\
\end{tblr}
\captionof{figure}{A video sequence of the clean ground truth with absolute timestamps and visibility labels. Two cars slow down, and the visibility labels are changed from 1.0 (green) to 0.0 (red). Bounding boxes with different colors correspond to different visibility (mobility).}
\label{tab:visualization_visibility}
\end{table}

\section{Implementation Details}
Our TEDNet is built on top of PermaTrack where its original modules and hyperparameters are kept unchanged. Instead of training the model on sequences of length 17 with a batch size 16 in PermaTrack, TEDNet is trained on sequences of length 4 with a batch size 9. Each 60-second chunk is split into a sequence of length 1200, i.e., each time window contains 50 ms event data. The memory of our recurrent X3D and ConvGRU are reset at the end of each batch of chunks. The recurrent X3D has a feature dimension of 640 $\times$ 360 with 3 $\times$ 3 convolution kernels, and ConvGRU has a feature dimension of 160 $\times$ 90 with 7 $\times$ 7 convolution kernels. The learning rate scheduler is the same as the one used in PermaTrack where it is set to $1.25 \times 10^{-4}$ and is decreased by a factor of 10 every 7 epochs for 1 epoch and increased back to the original value. All models are optimized using the Adam optimizer on 2 Nvidia 3090 GPUs with 100 epochs.

\begin{table}[!ht]
\centering
\begin{tblr}{
  row{15} = {font=\footnotesize, c}, 
  cell{1}{1} = {c=5}{},
  cell{3}{1} = {c=5}{},
  cell{5}{1} = {c=5}{},
  cell{7}{1} = {c=5}{},
  cell{9}{2} = {c=5}{},
  cell{11}{1} = {c=5}{},
  cell{13}{1} = {c=5}{},
  rows = {rowsep=0pt},
  columns = {colsep=1pt},
}
\centering
Ground Truth &  &  &  &  &  \\
\begin{subfigure}{\textwidthC}
    \includegraphics[width=\linewidth]{\ImgF{gt}{1063}}
\end{subfigure} &
\begin{subfigure}{\textwidthC}
    \includegraphics[width=\linewidth]{\ImgF{gt}{1102}}
\end{subfigure} & 
\begin{subfigure}{\textwidthC}
    \includegraphics[width=\linewidth]{\ImgF{gt}{1107}}
\end{subfigure} & 
\begin{subfigure}{\textwidthC}
    \includegraphics[width=\linewidth]{\ImgF{gt}{1113}}
\end{subfigure} &
\begin{subfigure}{\textwidthC}
    \includegraphics[width=\linewidth]{\ImgF{gt}{1137}}
\end{subfigure} \\
RED 0.1 &  &  &  &  &  \\
\begin{subfigure}{\textwidthC}
    \includegraphics[width=\linewidth]{\ImgF{RED_0.1_model_31}{1063}}
\end{subfigure} &
\begin{subfigure}{\textwidthC}
    \includegraphics[width=\linewidth]{\ImgF{RED_0.1_model_31}{1102}}
\end{subfigure} & 
\begin{subfigure}{\textwidthC}
    \includegraphics[width=\linewidth]{\ImgF{RED_0.1_model_31}{1107}}
\end{subfigure} & 
\begin{subfigure}{\textwidthC}
    \includegraphics[width=\linewidth]{\ImgF{RED_0.1_model_31}{1113}}
\end{subfigure} &
\begin{subfigure}{\textwidthC}
    \includegraphics[width=\linewidth]{\ImgF{RED_0.1_model_31}{1137}}
\end{subfigure} \\
RED 0.4 &  &  &  &  &  \\
\begin{subfigure}{\textwidthC}
    \includegraphics[width=\linewidth]{\ImgF{RED_0.4_model_31}{1063}}
\end{subfigure} &
\begin{subfigure}{\textwidthC}
    \includegraphics[width=\linewidth]{\ImgF{RED_0.4_model_31}{1102}}
\end{subfigure} & 
\begin{subfigure}{\textwidthC}
    \includegraphics[width=\linewidth]{\ImgF{RED_0.4_model_31}{1107}}
\end{subfigure} & 
\begin{subfigure}{\textwidthC}
    \includegraphics[width=\linewidth]{\ImgF{RED_0.4_model_31}{1113}}
\end{subfigure} &
\begin{subfigure}{\textwidthC}
    \includegraphics[width=\linewidth]{\ImgF{RED_0.4_model_31}{1137}}
\end{subfigure} \\
PermaTrack 0.1 &  &  &  &  &  \\
\begin{subfigure}{\textwidthC}
    \includegraphics[width=\linewidth]{\ImgF{PermaTrack_0.1_model_54}{1063}}
\end{subfigure} &
\begin{subfigure}{\textwidthC}
    \includegraphics[width=\linewidth]{\ImgF{PermaTrack_0.1_model_54}{1102}}
\end{subfigure} & 
\begin{subfigure}{\textwidthC}
    \includegraphics[width=\linewidth]{\ImgF{PermaTrack_0.1_model_54}{1107}}
\end{subfigure} & 
\begin{subfigure}{\textwidthC}
    \includegraphics[width=\linewidth]{\ImgF{PermaTrack_0.1_model_54}{1113}}
\end{subfigure} &
\begin{subfigure}{\textwidthC}
    \includegraphics[width=\linewidth]{\ImgF{PermaTrack_0.1_model_54}{1137}}
\end{subfigure} \\
PermaTrack 0.4 &  &  &  &  &  \\
\begin{subfigure}{\textwidthC}
    \includegraphics[width=\linewidth]{\ImgF{PermaTrack_0.4_model_54}{1063}}
\end{subfigure} &
\begin{subfigure}{\textwidthC}
    \includegraphics[width=\linewidth]{\ImgF{PermaTrack_0.4_model_54}{1102}}
\end{subfigure} & 
\begin{subfigure}{\textwidthC}
    \includegraphics[width=\linewidth]{\ImgF{PermaTrack_0.4_model_54}{1107}}
\end{subfigure} & 
\begin{subfigure}{\textwidthC}
    \includegraphics[width=\linewidth]{\ImgF{PermaTrack_0.4_model_54}{1113}}
\end{subfigure} &
\begin{subfigure}{\textwidthC}
    \includegraphics[width=\linewidth]{\ImgF{PermaTrack_0.4_model_54}{1137}}
\end{subfigure} \\
TEDNet 0.1 &  &  &  &  &  \\
\begin{subfigure}{\textwidthC}
    \includegraphics[width=\linewidth]{\ImgF{TEDNet_0.1_model_96}{1063}}
\end{subfigure} &
\begin{subfigure}{\textwidthC}
    \includegraphics[width=\linewidth]{\ImgF{TEDNet_0.1_model_96}{1102}}
\end{subfigure} & 
\begin{subfigure}{\textwidthC}
    \includegraphics[width=\linewidth]{\ImgF{TEDNet_0.1_model_96}{1107}}
\end{subfigure} & 
\begin{subfigure}{\textwidthC}
    \includegraphics[width=\linewidth]{\ImgF{TEDNet_0.1_model_96}{1113}}
\end{subfigure} &
\begin{subfigure}{\textwidthC}
    \includegraphics[width=\linewidth]{\ImgF{TEDNet_0.1_model_96}{1137}}
\end{subfigure} \\
TEDNet 0.4 &  &  &  &  &  \\
\begin{subfigure}{\textwidthC}
    \includegraphics[width=\linewidth]{\ImgF{TEDNet_0.4_model_96}{1063}}
\end{subfigure} &
\begin{subfigure}{\textwidthC}
    \includegraphics[width=\linewidth]{\ImgF{TEDNet_0.4_model_96}{1102}}
\end{subfigure} & 
\begin{subfigure}{\textwidthC}
    \includegraphics[width=\linewidth]{\ImgF{TEDNet_0.4_model_96}{1107}}
\end{subfigure} & 
\begin{subfigure}{\textwidthC}
    \includegraphics[width=\linewidth]{\ImgF{TEDNet_0.4_model_96}{1113}}
\end{subfigure} &
\begin{subfigure}{\textwidthC}
    \includegraphics[width=\linewidth]{\ImgF{TEDNet_0.4_model_96}{1137}}
\end{subfigure} \\
(a) 53.15 s & (b) 55.10 s & (c) 55.35 s & (d) 55.65 s & (e) 56.85 s \\
$\xrightarrow{\hspace*{2.2cm}}$ & time \\
\end{tblr}

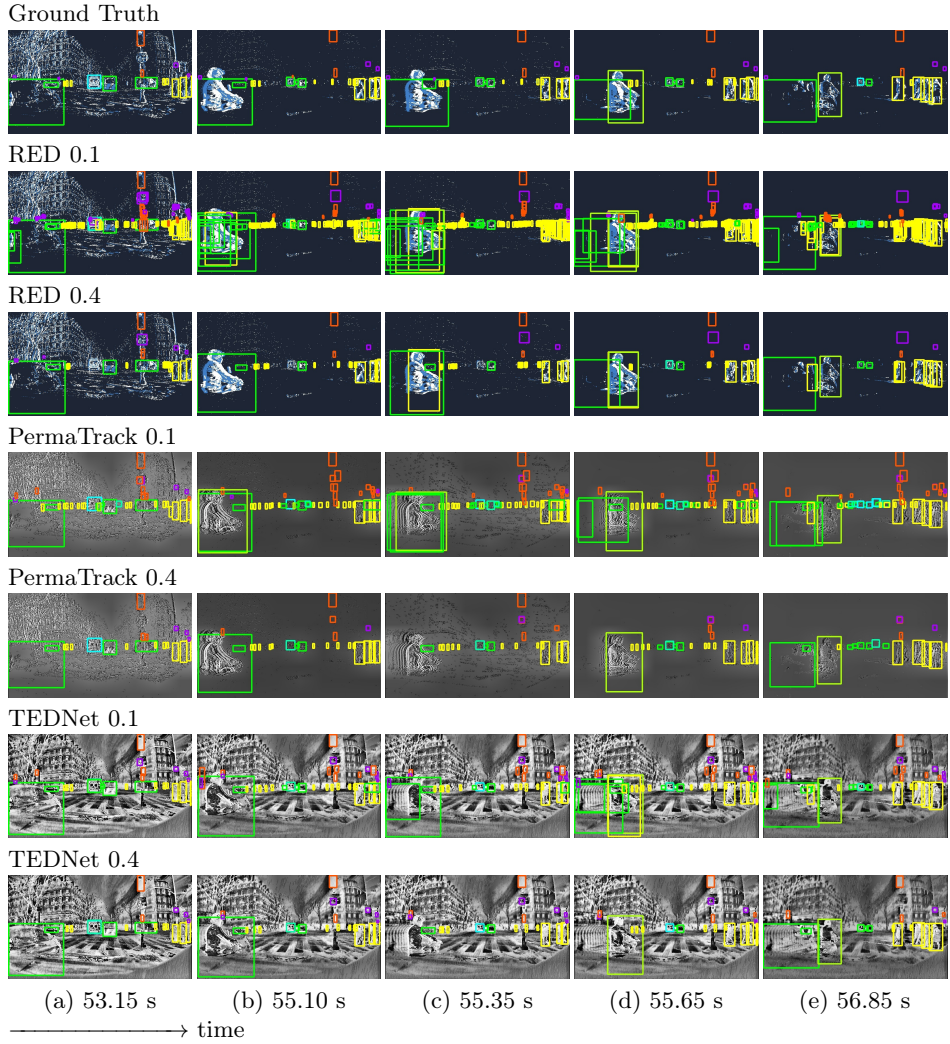
\captionof{figure}{A video sequence of the failure case with absolute timestamps, ground truth labels, and some detection results with different confidence thresholds. One car on the left remains still for a while, and one motorcyclist passes between the still car and the camera. The values after model names are the thresholds where the bounding boxes with confidence scores lower than those thresholds are filtered out. Bounding boxes with different colors correspond to different categories. }
\label{tab:visualization_failure}
\end{table}

\section{Visualization of Failure Cases}
We visualize the failure case in \cref{tab:visualization_failure} by a video with one car on the left remaining still for a while and one motorcyclist passing between the still car and the camera. This scenario is different from the one in \cref{sec:Visualization} where no intervention between the still objects and the camera. The root cause of the failure case in \cref{tab:visualization_failure} comes from the degradation of confidence scores when the intervention occurs between the still objects and the camera. Meanwhile, the \emph{D3D} in our TEDNet causes the ghosting effect of the motorcyclist in the 3rd and 4th frames of our TEDNet where the wrong and low-confidence-score bounding box is produced to detect the residual features. As a result, neither PermaTrack nor our TEDNet works well when the intervention occurs between the still objects and the camera. Namely, even though we regard those still and invisible objects as pseudo-occlusions, such that our TEDNet can track them and keep the bounding boxes, our TEDNet has weak capabilities to track real-occlusion with a high confidence score. However, the prior work RED can maintain the high-confidence-score bounding box even if the intervention occurs due to its implicit memory architecture and no feature missing for a very long time, which is different from \cref{sec:Visualization}.

\section{Supplementary Video}
We provide a supplementary video, where the images from the main paper are captured, to demonstrate the problem of prior works and the result of our TEDNet. There are two cars remaining still for a very long time in the video, i.e., the left car remains still from 22.5 to 55.9 seconds and the right car remains still from 22.5 to 52.7 seconds. \cref{tab:MissingBbox} shows the duration of missing bounding boxes of RED, DMANet, and HMNet in the video for the left and the right car, respectively. As mentioned in the main paper, the missing bounding boxes last 17.9 seconds in RED, 32.4 seconds in DMANet, and interweavingly in HMNet. In addition, the video from 37.6 to 38.6 seconds shows that the real occluded objects are tracked by RVT, CenterTrack, and PermaTrack, incurring many false positives. Instead, our TEDNet discards those real occluded objects achieving fewer false positives and higher mAP performance. The video is publicly available at \url{https://youtu.be/JCCZcmZ2oXU}.

\begin{table}[!b]
\centering
\caption{The duration of missing bounding boxes in the supplementary video for the left and the right car, respectively.}
\label{tab:MissingBbox}
\begin{tabular}{>{\centering\arraybackslash}p{2.8cm} >{\centering\arraybackslash}p{2.8cm} >{\centering\arraybackslash}p{2.8cm}}
\toprule
Methods & Left Car          & Right Car         \\
\midrule
RED    & 38.4 $\sim\ $56.3 s & no missing bbox   \\
\midrule
DMANet & 23.7 $\sim\ $56.1 s & 25.6 $\sim\ $52.9 s \\
\midrule
HMNet &
  \begin{tabular}[c]{@{}l@{}}30.9 $\sim\ $31.0 s\\ 34.1 $\sim\ $34.8 s\\ 36.2 $\sim\ $36.8 s\\ 46.1 $\sim\ $46.4 s\\ 48.2 $\sim\ $49.0 s\\ 49.2 $\sim\ $49.7 s\end{tabular} &
  no missing bbox \\
\midrule
TEDNet    & no missing bbox & no missing bbox   \\
\bottomrule
\end{tabular}
\end{table}

\section{Quantitative Analysis}
We compare our TEDNet to the state-of-the-art event-based object detectors on \emph{clean GT} to demonstrate why our TEDNet can improve the mAP significantly. The comparison can be divided into two parts, including fine-grained and category-level comparison. The confidence score is set to 0.4 for all of the models.

\subsection{Fine-grained Comparison}

\begin{table}[!t]
  \caption{Fine-grained comparison with the state-of-the-art event-based object detectors on \textbf{clean GT}. The number of objects shown in the table is the mantissa of scientific notation with base 10 and exponent 3. (GT: Total Ground Truth / DT: Total Detection / TP: True Positive / FP: False Positive / FN: False Negative)}
  \label{tab:FineGrainCompare}
  \centering
  \begin{tabular}{>{\centering\arraybackslash}p{2.8cm} >{\centering\arraybackslash}p{1.2cm} >{\centering\arraybackslash}p{2cm} >{\centering\arraybackslash}p{2cm} >{\centering\arraybackslash}p{1.5cm}}
    \toprule
    Methods & RED & CenterTrack & PermaTrack & TEDNet \\ 
    \midrule
    GT & 1220 & 1220 & 1220 & 1220 \\
    DT & 1466 & \textbf{1134} & 1307 & \underline{1218} \\
    TP & 829 & 836 & \underline{898} & \textbf{930} \\
    FP(wrong id)\textsuperscript{*} & 29 & \textbf{24} & 34 & \underline{25} \\
    FP(wrong bbox)\textsuperscript{\dag} & 608 & \underline{274} & 376 & \textbf{263} \\
    FP & 637 & \underline{298} & 409 & \textbf{288} \\
    FN  & 362 & 359 & \underline{288} & \textbf{264} \\
    \midrule
    Precision  & 56.6 & \underline{73.7} & 68.7 & \textbf{76.3} \\
    Recall  & 68.0 & 68.6 & \underline{73.6} & \textbf{76.3} \\
    \bottomrule
    \multicolumn{5}{l}{The bold and the underline indicate the best and the second-best.} \\
    \multicolumn{5}{l}{* Wrong id means the bounding box is correctly detected with the} \\
    \multicolumn{5}{l}{\ \ \ wrong category identity.} \\
    \multicolumn{5}{l}{\dag\ Wrong bbox means the bounding box is incorrectly detected.}
  \end{tabular}
\end{table}

We calculate the number of ground truths (GT), detections (DT), true positives (TP), false positives (FP), and false negatives (FN) with a predefined confidence score 0.4 as shown in \cref{tab:FineGrainCompare}. All of the numbers on our TEDNet are either the best or the second-best, where our TEDNet not only increases TPs and decreases FNs, but also decreases FPs significantly and decreases DTs, making it more robust than other models. In summary, our TEDNet achieves the highest precision and recall.

\subsection{Category-level Comparison}
We conduct category-level analysis to demonstrate the effectiveness of our TEDNet on still objects in \cref{tab:StillObj} and moving objects in \cref{tab:MoveObj}. The reason why the objects are split into still and moving objects in the category-level comparison is to prove our intuition of regarding still objects as pseudo-occlusion and exploiting tracking through occlusion to keep object permanence. For the still objects in \cref{tab:StillObj}, our TEDNet increases the TPs of all the categories by a significant amount, demonstrating the efficacy of tracking through pseudo-occlusion. For moving objects in \cref{tab:MoveObj}, our TEDNet performs either the best or the second-best for every category, and it performs the best in total. In summary, our TEDNet can not only improve the number of TPs on still objects but also the number of TPs on moving objects, making it achieve state-of-the-art performance.

\begin{table}[!t]
\centering
\caption{Category-level comparison with the state-of-the-art event-based object detectors on \textbf{still objects} of \textbf{clean GT}. The number of detected objects (TP) shown in the table is the mantissa of scientific notation with base 10 and exponent 3.}
\label{tab:StillObj}
\begin{tabular}{>{\centering\arraybackslash}p{2cm} >{\centering\arraybackslash}p{1.2cm} >{\centering\arraybackslash}p{1.2cm} >{\centering\arraybackslash}p{2cm} >{\centering\arraybackslash}p{2cm} >{\centering\arraybackslash}p{1.5cm}}
\toprule
Categories & GT & RED & CenterTrack & PermaTrack & TEDNet \\
\midrule
pedestrian    & 27  & 11.9 & 12.3  & \underline{15}  & \textbf{18} \\
two wheeler   & 7.9 & 3.6  & 4.6   & \underline{5.7} & \textbf{6.0} \\
car           & 90  & 50   & 70    & \underline{75}  & \textbf{79} \\
truck         & 5.0 & 1.6  & 2.9   & \underline{3.1} & \textbf{3.2} \\
bus           & 4.3 & 1.5  & \underline{1.82}  & 1.7 & \textbf{1.83} \\
traffic sign  & 19  & 3    & 8     & \underline{10}  & \textbf{11} \\
traffic light & 41  & 12   & 21    & \underline{25}  & \textbf{27} \\
total         & 195 & 84   & 121   & \underline{134} & \textbf{146} \\
\bottomrule
\multicolumn{6}{l}{The bold and the underline indicate the best and the second-best.}
\end{tabular}
\end{table}

\begin{table}[!t]
\centering
\caption{Category-level comparison with the state-of-the-art event-based object detectors on \textbf{moving objects} of \textbf{clean GT}. The number of detected objects (TP) shown in the table is the mantissa of scientific notation with base 10 and exponent 3.}
\label{tab:MoveObj}
\begin{tabular}{>{\centering\arraybackslash}p{2cm} >{\centering\arraybackslash}p{1.2cm} >{\centering\arraybackslash}p{1.2cm} >{\centering\arraybackslash}p{2cm} >{\centering\arraybackslash}p{2cm} >{\centering\arraybackslash}p{1.5cm}}
\toprule
Categories & GT & RED & CenterTrack & PermaTrack & TEDNet \\
\midrule
pedestrian    & 286 & \underline{203} & 178 & 200 & \textbf{213} \\
two wheeler   & 41  & 26 & 28.9 & \textbf{30} & \underline{29.3} \\
car           & 480 & 405 & 397 & \underline{409} & \textbf{410} \\
truck         & 56  & 32 & 35 & \underline{40} & \textbf{42} \\
bus           & 19  & \underline{10} & 9 & 8 & \textbf{12} \\
traffic sign  & 78  & 40.6 & \underline{41.2} & 40.9 & \textbf{44} \\
traffic light & 63  & 29 & 28 & \textbf{37} & \underline{36} \\
total         & 1025 & 746 & 716  & \underline{764} & \textbf{785} \\
\bottomrule
\multicolumn{6}{l}{The bold and the underline indicate the best and the second-best.}
\end{tabular}
\end{table}

\end{document}